\documentclass{Interspeech}

\interspeechcameraready 

\title{Triadic Multi-party Voice Activity Projection for Turn-taking \texorpdfstring{\\}{}in Spoken Dialogue Systems}

\author{Mikey}{Elmers}
\author{Koji}{Inoue}
\author{Divesh}{Lala}
\author{Tatsuya}{Kawahara}

\affiliation{Graduate School of Informatics}{Kyoto University}{Japan}
\email{[elmers, inoue, lala, kawahara]@sap.ist.i.kyoto-u.ac.jp}

\usepackage{comment}

\begin{document}

\maketitle

\begin{abstract}

Turn-taking is a fundamental component of spoken dialogue, however conventional studies mostly involve dyadic settings.
This work focuses on applying voice activity projection (VAP) to predict upcoming turn-taking in triadic multi-party scenarios.
The goal of VAP models is to predict the future voice activity for each speaker utilizing only acoustic data. 
This is the first study to extend VAP into triadic conversation.
We trained multiple models on a Japanese triadic dataset where participants discussed a variety of topics.
We found that the VAP trained on triadic conversation outperformed the baseline for all models but that the type of conversation affected the accuracy.
This study establishes that VAP can be used for turn-taking in triadic dialogue scenarios.
Future work will incorporate this triadic VAP turn-taking model into spoken dialogue systems.
    
\end{abstract}

\section{Introduction}\label{sec:intro}

Turn-taking is essential to human-human conversation determining who should talk and when \cite{sacks1974}, with successful communication requiring synchronization between speakers and addressees \cite{clark2002speechcomm}.
Turn-taking exhibits language-independent characteristics, such as a focus on fewer silences and less overlapping speech \cite{stivers2009universals}.
However, overlapping speech is largely influenced by both the type of conversation and the number of participants involved.
A fundamental problem within turn-taking research is that we find short delays during turn-taking with gaps between \SI{100}{} $-$ \SI{300}{\milli\second} even though we find language production latencies between \SI{600}{} $-$ \SI{1500}{\milli\second} \cite{bates2003timing}.
This suggests that humans prepare the next utterance even before the current utterance is completed by predicting turn-taking.
In fact, long gaps of silence can cause issues in conversation.
For example, across a variety of languages (e.g., American English, Italian, and Japanese) long gaps of silence reduced ratings for both requests and assessments \cite{roberts2011discourse}.

Spoken dialogue systems (SDSs) have become commonplace in everyday life but still struggle with a number of issues \cite{ward2016aimag}, including turn-taking.
Turn-taking in SDSs has traditionally been inflexible \cite{ward2010interspeech}, using silence thresholds for determining when the system should take the turn \cite{skantze2014speechcomm}.
These silence thresholds usually range between \SI{500}{} $-$ \SI{1000}{\milli\second} \cite{ferrer2002icslp}, incurring trade-offs such as: 1) delays for the system response, and 2) interruptions for the speaker who has not finished their turn.
The poor turn-taking quality of SDSs results in users adopting an unnatural turn-taking style, including false starts before the system begins speaking.
In order to improve the performance of SDSs, it is necessary to develop turn-taking systems that reflect human-human conversation, since response timing is an important contributor for subjective evaluations of naturalness \cite{itoh2009interspeech}.

\textbf{Voice Activity Projection} (\textbf{VAP}) \cite{ekstedt22interspeech} has been widely used as a turn-taking model to predict joint future voice activity using only acoustic information.
VAP has been used in many scenarios including: real-time continuous turn-taking\footnote{\url{https://arxiv.org/abs/2401.04868}}, multilingual \cite{inoue2024lreccoling} and multi-modal \cite{onishi2023hai} turn-taking, filled pause analysis \cite{binger2023icphs}, backchannel prediction \cite{liermann2023emnlp}, and incremental response generation \cite{chiba2025acl}.
However, research using VAP has \textit{exclusively} focused on dyadic conversation.
The goal of this work is to extend VAP to a multi-party setting, specifically three-person (i.e., triadic) conversations.

The major contribution of this paper is the development and analysis of triadic VAP models.
This is the first study in which VAP has been used for triadic multi-party turn-taking.
The methods, including data, pre-processing, model architecture, and model training, are presented in Section~\ref{sec:method}, the results in Section~\ref{sec:result}, a general discussion and the limitations of this work in Section~\ref{sec:discussion}, and a summary and conclusion in Section~\ref{sec:conclusion}.

\section{Methods}\label{sec:method}
\subsection{Voice Activity Projection}
\begin{figure}[t]
  \centering
  \includegraphics[width=0.65\linewidth]{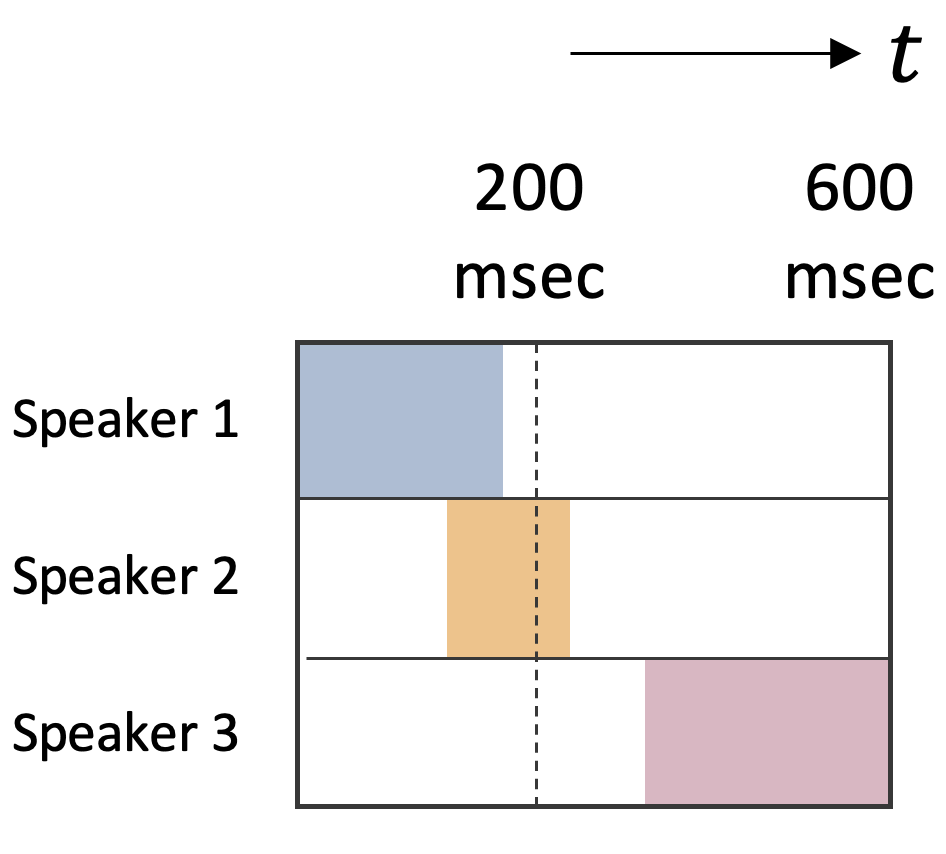}
  \caption{Discretized bins for triadic VAP model. The dashed vertical line demarcates the bins and the segments indicate speaker voice activity.}
  \label{fig:vap_states}
\end{figure}

VAP predicts future voice activity for all speakers jointly using only acoustic information.
The voice activity is coded as a binary (i.e., speaking or not speaking) sequentially throughout the conversation.
Traditionally, dyadic VAP models have predicted a state window of \SI{2}{\second} of future voice activity discretized into four sub-state bins per speaker with durations of \SI{0.2}{\second}, \SI{0.4}{\second}, \SI{0.6}{\second}, and \SI{0.8}{\second}, resulting in eight sub-state bins (i.e., 2 speakers * 4 bins), or $2^{8} = 256$ possible states.
Future voice activity predictions become more complicated as we add additional speakers because as the number of participants increases the number of possible states increases as well.
Therefore, we opted to evaluate our triadic model with two sub-states bins for each speaker resulting in six sub-state bins (i.e., 3 speakers * 2 bins), or $2^{6} = 64$ possible states.
Figure~\ref{fig:vap_states} shows an example of our triadic VAP model.
We chose to maintain the sub-state bins with durations of \SI{0.2}{\second} and \SI{0.4}{\second} rather than extending our two sub-state bins to cover the entire \SI{2}{\second} that traditional dyadic models cover.
A bin is determined as active if a majority of the bin contains voice activity.
The bins are averaged using a probability distribution to determine overall voice activity.
The discrete states are used as labels during training.
For more information regarding VAP states, see \cite{ekstedtphd}.

\subsection{Dataset}
\begin{figure}[t]
  \centering
  \includegraphics[width=\linewidth]{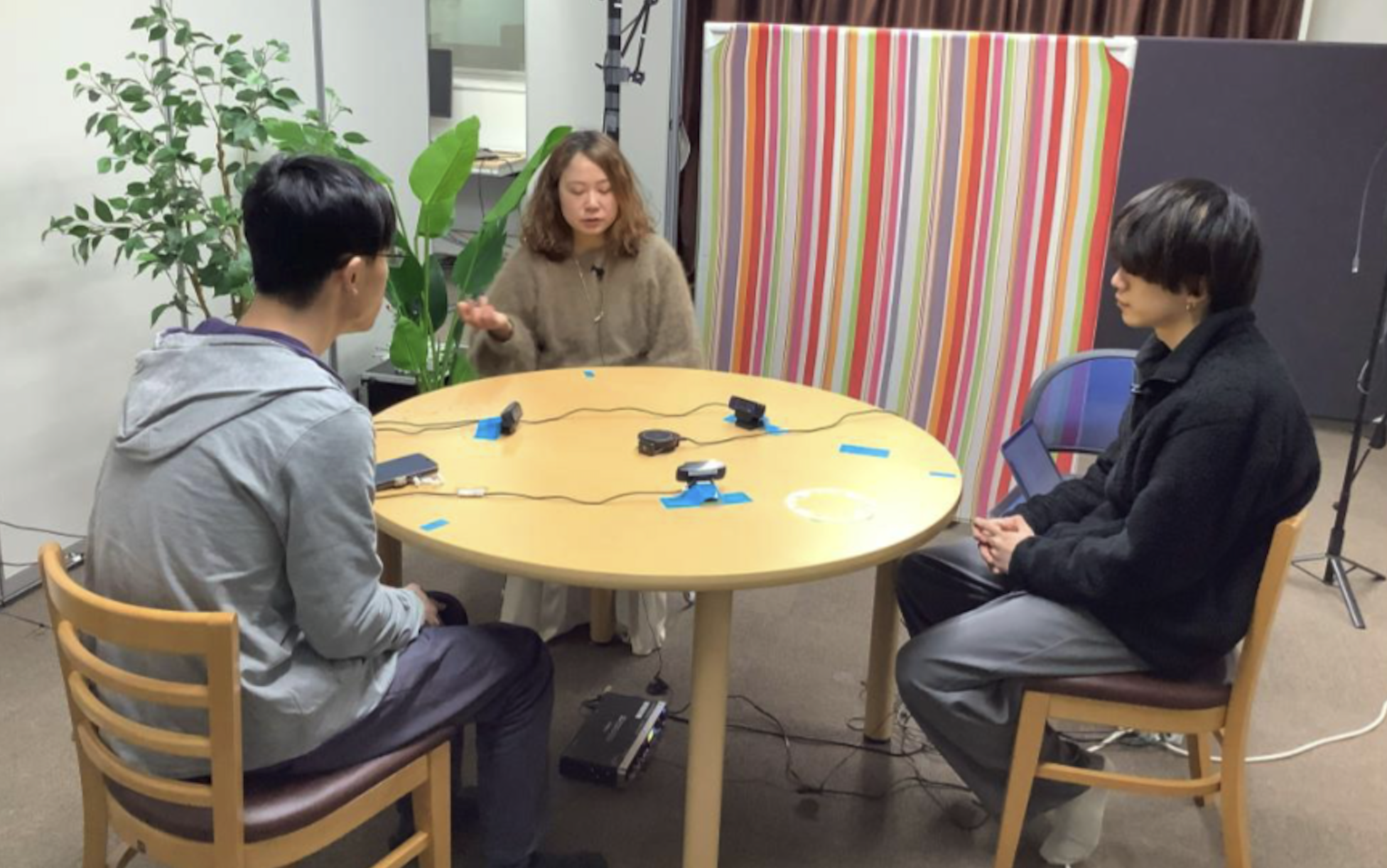}
  \caption{Example session setup from TEIDAN corpus.}
  \label{fig:teidan}
\end{figure}

The VAP model for this experiment was trained using our TEIDAN corpus.
We constructed the TEIDAN corpus as a multi-modal multi-party dialogue corpus of triadic discussions because we were unable to find other datasets that contained spontaneous triadic conversations with separated audio channels for each speaker.
The participants sat in a circle as seen in Figure~\ref{fig:teidan}.
Each participant had a separate pin microphone, and a microphone array was placed at the center of the table.
Separate cameras individually recorded each participant.
The discussions were conducted entirely in Japanese.
The corpus consists of both spontaneous discussions and attentive listening.
Participants were instructed to engage in free discussion in all scenarios with no forced ending point.
Half of the participants in the spontaneous triads discussed the following three topics: 1) possible alternative capital cities for Japan, 2) items necessary for survival on a deserted island, and 3) potential travel locations for a weekend trip.
The other half of participants in the spontaneous triads discussed an alternative set of three topics: 1) ranking money, familiy, or experiences based on personal importance, 2) choosing to visit a city, mountain, or beach for a fun trip, and 3) traveling to Tokyo by bullet train, plane, or car.
The same triad discussed all topics resulting in three sessions per triad group.
The attentive listening triads involved a primary speaker who spoke about whatever topic they wanted while the other two participants responded as active listeners.
In the attentive listening triads all three participants performed as both the primary speaker and active listener roles.
We collected data from eighteen unique triads (twelve spontaneous and six attentive listening) resulting in a total of fifty-four discussions.
Table~\ref{tab:teidan_duration} shows the duration values for the corpus.
Manual voice activity annotations were made for all speakers for training the VAP model.

\begin{table}[th]
  \caption{Duration information for the TEIDAN dataset. The mean (Avg.) and total (Total) duration for the spontaneous (SP), attentive listening (ATT), and combined (BOTH) datasets are provided. Duration values are measured in minutes.}
  \label{tab:teidan_duration}
  \centering
  \begin{tabular}{c c c}
    \toprule
    {\textbf{Category}} & {\textbf{Avg.}} & {\textbf{Total}} \\
    \midrule
    SP & 6.20 & 223.36~~~ \\
    ATT & 3.58 & 64.43~~~ \\
    BOTH & 5.33 & 287.79~~~ \\
    \bottomrule
  \end{tabular}

\end{table}

We evaluated the amount of overlapping speech present in the TEIDAN corpus.
This is important because dyadic and triadic speech can differ in the number of overlapping speakers.
This problem is further intensified by the spontaneous speech found within our corpus.
As previously mentioned, our dataset included manual annotations of voice activity for each speaker.
We calculated overlapping segments of speech by locating the voice activity labels for each participant and evaluating the following categories: mutual silence (i.e., no voice activity from any speaker), single speaker (i.e, voice activity from a single member of the triad), two speakers (i.e., simultaneous voice activity from two members of the triad), and three speakers (i.e., simultaneous voice activity from all members of the triad).
Table~\ref{tab:teidan_overlap_full} contains the overlapping speech information for the TEIDAN corpus.
On average, overlapping speech accounted for 31.98\% of the total duration in the spontaneous data, and 21.17\% in the attentive listening data.
The presence of overlapping speech in the attentive listening data can be largely explained as backchanneling.

\begin{table}[th]
  \caption{Overlapping speech information for the TEIDAN dataset. The mean percent (out of 100\%) is provided for mutual silence, a single speaker, two speakers, and three speakers for spontaneous (SP), attentive listening (ATT), and combined (BOTH).}
  \label{tab:teidan_overlap_full}
  \centering
  \begin{tabular}{c c c c c}
  \toprule
  \textbf{} & \textbf{Silence} & \textbf{Single} & \textbf{Double} & \textbf{Triple} \\
  \midrule
  SP   & 12.02\% & 56.01\% & 24.24\% & 7.74\% \\
  ATT  & 15.47\% & 63.36\% & 15.68\% & 5.49\% \\
  BOTH & 13.11\% & 58.82\% & 21.31\% & 6.76\% \\
  \bottomrule
\end{tabular}

\end{table}

\subsection{Pre-Processing}
The VAP model for this experiment is trained using a three-channel audio (i.e., one channel for each speaker) with a \SI{16}{\kilo\hertz} sampling rate, and a 16-bit bit depth.
We did not incorporate any form of diarization since each participant had their own pin microphone.
The VAP model evaluates the audio and voice activity information in \SI{20}{\second} intervals.
We included the start time and end time for each voice activity segment for all three speakers for each \SI{20}{\second} interval.
Since the model predicts the future \SI{600}{\milli\second} of voice activity, we included voice activity information up to \SI{20.6}{\second} for each interval.
The VAP model was trained on the raw audio and voice activity information for each \SI{20}{\second} interval.

\subsection{Model Architecture}
\begin{figure}[t]
  \centering
  \includegraphics[width=\linewidth]{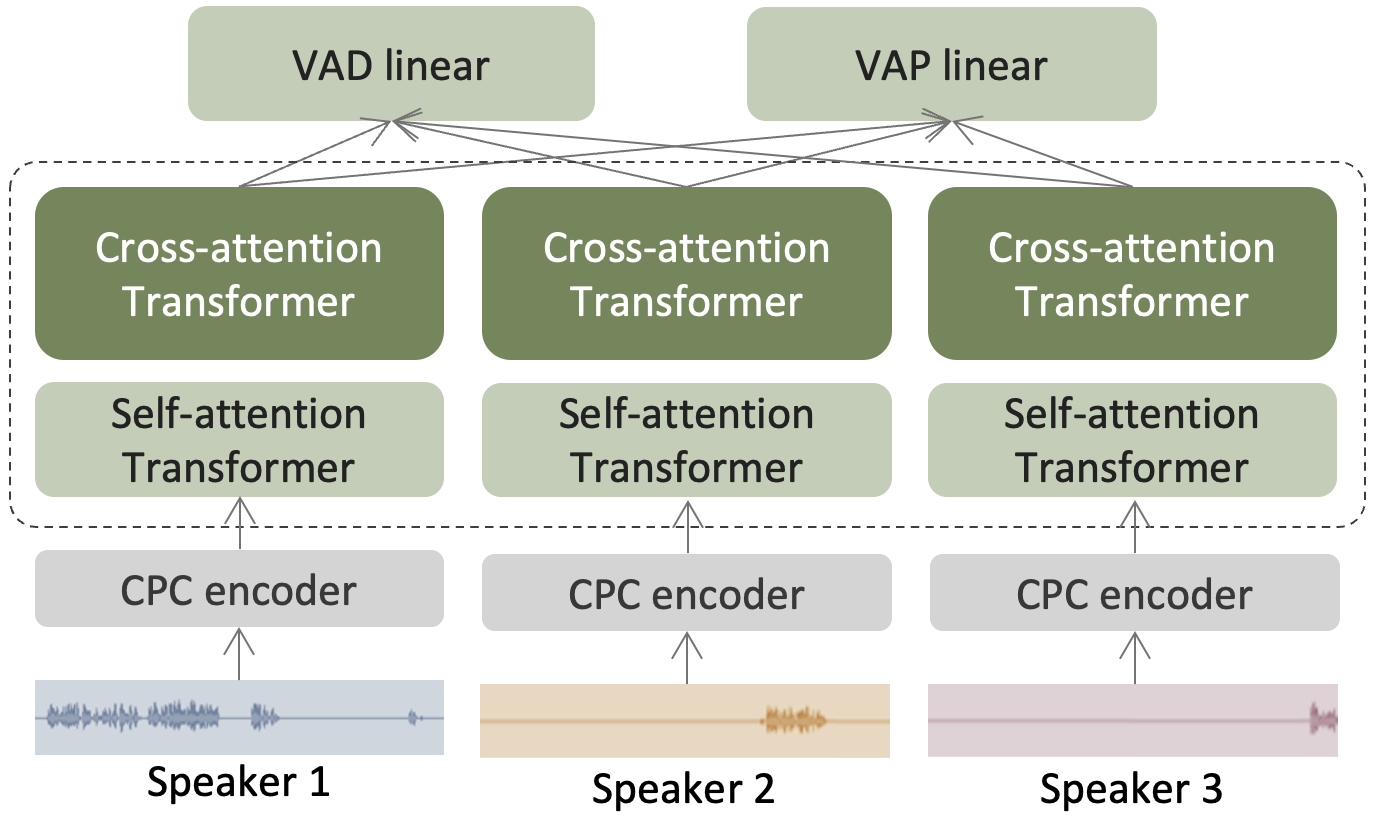}
  \caption{Triadic VAP model architecture.}
  \label{fig:vap_architecture}
\end{figure}

The triadic VAP model architecture is shown in Figure~\ref{fig:vap_architecture}.
Similar to other VAP studies, we used a pre-trained contrastive predictive coding (CPC) \footnote{\url{https://arxiv.org/abs/1807.03748}} encoder to extract features from the raw audio. 
The CPC was pre-trained on the LibriSpeech dataset \cite{librispeech} using the methods described in \cite{riviere2020icassp}.
The CPC output is fed to a one-layer self-attention Transformer \cite{vaswani2017} that independently evaluates each audio channel.
The outputs from the self-attention Transformer are inputted into a 64-dimension three-layer multi-channel cross-attention Transformer which encodes interactive information from the three channels.
Next, we concatenate the outputs from the Transformers and perform multitask learning.
The primary task is VAP which predicts future voice activity of the three participants, encoded in Figure~\ref{fig:vap_states}.
The second task is voice activity detection (VAD) which detects current voice activity for the three participants to help stabilize the VAP training \cite{inoue2024lreccoling}.
Overall, the triadic VAP model architecture is largely the same as the architecture used for dyadic VAP.
The dyadic cross-attention was extended to a multi-channel cross attention Transformer designed to handle triadic data.

\subsection{Model Training}
The VAP model uses the previous voice activity of \SI{20}{\second} and the accompanying waveform as input to predict the future voice activity of \SI{0.6}{\second} using a \SI{50}{\hertz} frame rate (i.e., frame predictions are made every \SI{20}{\milli\second}).
We split the data using 80\% for training, 10\% for validation, and 10\% for testing with a random seed set for reproducibility.
The data was split so that data from a specific triad was not used in multiple splits.
The model was trained for 10 epochs with a batch size of 8 and a weight decay of 0.001.
We used the AdamW optimizer with a learning rate of \SI{3.64e-4}{}.
Cross-entropy loss was used as the loss function, comparing ground truth labels and predicted voice activity during training.
The test set was evaluated using the model that produced the smallest loss on the validation split.
The model was implemented using PyTorch \cite{pytorch} and based on code from previous work \cite{inoue2024lreccoling}.

\section{Results}\label{sec:result}
\subsection{Loss}
We evaluated three models: 1) a model trained only on the spontaneous speech from the TEIDAN corpus, 2) a model trained only on the attentive listening part of the TEIDAN corpus, and 3) a model that was trained on both spontaneous and attentive listening.
Table~\ref{tab:model_loss} contains the test loss for the three models.
The model with both spontaneous speech and attentive listening performed best while the attentive listening only model performed the worst.

\begin{table}[th]
  \caption{Model test loss performance for only spontaneous (SP), only attentive listening (ATT), and combined (BOTH).}
  \label{tab:model_loss}
  \centering
  \begin{tabular}{c c}
    \toprule
    {\textbf{Model}} & {\textbf{Loss}} \\
    \midrule
    SP & 2.31~~~ \\
    ATT & 2.51~~~ \\
    BOTH & 2.09~~~ \\
    \bottomrule
  \end{tabular}

\end{table}

\subsection{Model Comparison}
We evaluated the VAP model by predicting the next speaker during periods of mutual silence.
First, we located segments of mutual silence (i.e., no voice activity from any of the speakers) that were at least \SI{100}{\milli\second} in duration.
Next, we determined the next speaker.
However, to prevent detecting backchannels or other short responses, we looked for speech that was at least \SI{1}{\second} in duration and made sure there was no other overlapping speech during that time.
We evaluated the model's expected probabilities from when the speaker begins up to \SI{600}{\milli\second} in the past (i.e., the maximal amount the model predicts).
Next, we aggregated the probabilities for the three speakers during this time.
Finally, we compared the speaker with the highest probability during the mutual silence to the upcoming speaker to determine if the model's prediction was accurate or not.
We compared against a last speaker baseline where the model predicted the last speaker as the current speaker.
Table~\ref{tab:model_acc} contains the accuracy for the three models.
All models performed better than the respective baseline.
The model trained only on spontaneous data had lower accuracy than those trained on attentive listening, as more participants speak simultaneously during discussions.
As expected, performance accuracy was much higher for the attentive listening data, where a single speaker predominately leads the conversation.

\begin{table}[th]
  \caption{Next speaker prediction comparing baseline and proposed method for only spontaneous (SP), only attentive listening (ATT), and combined (BOTH).}
  \label{tab:model_acc}
  \centering
  \begin{tabular}{c c c}
    \toprule
    {\textbf{Model}} & {\textbf{Baseline}} & {\textbf{Proposed}} \\
    \midrule
    SP & 55.56\% & 62.43\%~~~ \\
    ATT & 81.25\% & 87.50\%~~~ \\
    BOTH & 77.58\% & 86.83\%~~~ \\
    \bottomrule
  \end{tabular}

\end{table}

\subsection{Predictions vs. Ground Truth Example}
\begin{figure}[t]
  \centering
  \includegraphics[width=\linewidth]{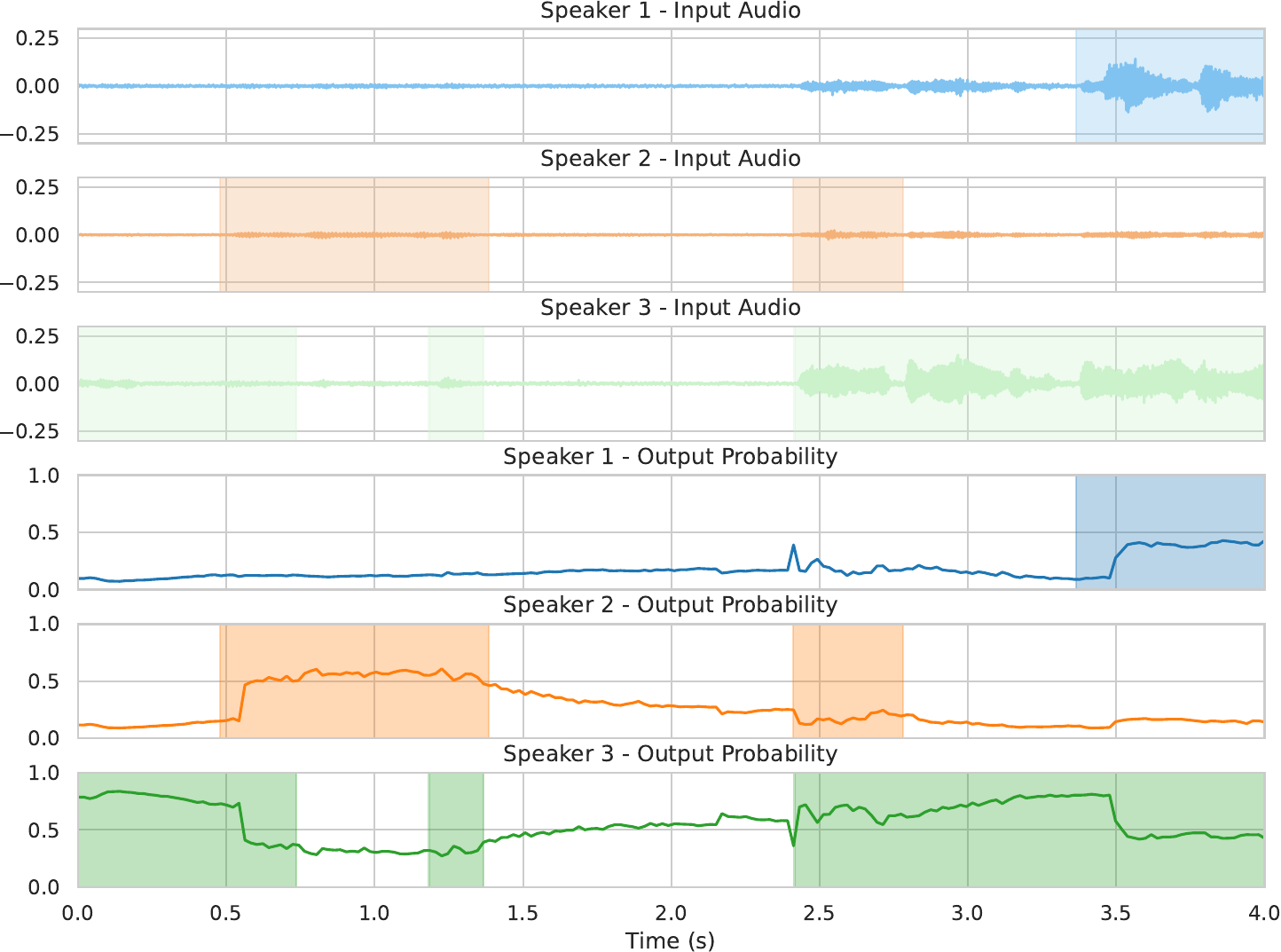}
  \caption{
  The top three plots show the ground truth waveforms (amplitude) for each of the three speakers, while the bottom three depict the model's predicted voice activity probabilities (0 to 1). Colors denote speaker: blue (1), orange (2), and green (3).}
  \label{fig:predictions}
\end{figure}

Figure~\ref{fig:predictions} provides an example comparing the model's predictions to ground truth labels during \SI{4}{\second} of dialogue.
At $\sim$ \SI{2.5}{\second}, speaker 3 (green) begins a turn while briefly overlapping with speaker 2 (orange).
The model indicates that speaker 3 has the highest output probability—correctly predicting the turn change—but underestimates speaker 2, illustrating the challenge of assigning probabilities during overlap, where predictions must exceed a threshold and are constrained to single-speaker outputs.
Importantly, the model accurately predicts the switch from speaker 2 to speaker 3 even though both speakers overlap before and after the mutual silence.

\section{Discussion}\label{sec:discussion}
In this work we evaluated a triadic VAP model and found that the model can accurately predict upcoming non-overlapping speech during periods of mutual silence.
However, we found overlapping speech to be especially common in our TEIDAN data.
This problem is magnified since our data uses spontaneous speech where speaker overlap is prevalent.
Overlapping speech found during conversation is influenced by a number of factors including: speaker familiarity (i.e., friends or strangers), language, gender, and conversation topic \cite{yuan2007icphs}.
Future work should consider how to predict turn-taking for overlapping speech especially if it remains quite common in other datasets.

The goal of this work was to develop a multi-party turn-taking system that can be implemented into embodied robots and virtual agents.
Therefore, future work will implement this triadic VAP model into a SDS to evaluate it in a variety of scenarios and see how the model performs in experimental conditions with subjective ratings.

There are some limitations to the approach used in this work.
Our current VAP implementation only used acoustic information.
However, content, syntax, prosody, paralanguage, and gesture are additional cues that contribute to turn-yielding \cite{duncan1972signals}.
Gravano et al. \cite{gravano2011computer} found that as the number of turn-yielding cues increases, so does the chance of a turn change.
To effectively model turn-taking requires a wealth of multi-modal information such as gaze \cite{onishi2023hai}, gesture, linguistic information, and speaker intention \cite{aldeneh2018icassp}.
Therefore, incorporating more multi-modal cues when modeling turn-taking improves the accuracy of identifying turn changes.
Incorporating visual and linguistic information will be especially important for predicting the next speaker.
In dyadic turn-taking it is implicitly assumed that when the current speaker has finished their turn the other speaker will initiate their turn.
However, multi-party turn-taking is more complex in that it is not immediately clear using only acoustic information who will take the turn after the current speaker finishes their turn.
This work was designed to find the limit of acoustic-only VAP for multi-party turn-taking.
However, it is essential to incorporate many cues from different modalities to accurately model multi-party turn-taking.

Another important area for consideration is the number of participants involved in the multi-party dialogue.
In the present study we focused exclusively on triadic multi-party conversation, however, future work should evaluate additional group sizes.
One issue inherent to the design of traditional VAP models is that, as the number of participants increases, the number of states increases significantly.
For example, if we maintained four bins per speaker (like other dyadic VAP studies), dyadic conversation results in $2^{8} = 256$ possible states, triadic conversation results in $2^{12} = 4096$ possible states, and tetradic conversation results in $2^{16} = 65536$ possible states.
The number of possible states quickly becomes unmanageable even before we reach group sizes of five to ten participants.
In addition to data sparsity issues, we encountered memory issues during training when we attempted to evaluate a triadic model with four sub-state bins.
Therefore, it is important to consider a reasonable amount of possible states to prevent data sparsity and training issues.
It is possible that the design of VAP might struggle to handle groups larger than four or five participants, and that alternative methods might be necessary for predicting turn-taking for larger group sizes. 
When designing the triadic VAP architecture we chose the multi-channel cross-attention Transformer to be scalable for larger group sizes.
Therefore, in future work we plan to evaluate group sizes larger than three-person conversations to find the limits of the VAP methodology with respect to group size.

Lastly, the TEIDAN dataset is a new corpus that is currently in development, continuing to expand, and will be released in the future.
Previous VAP models were evaluated on large dyadic datasets that are not available for triadic conversation.
In this study, we evaluated our triadic VAP model exclusively on the TEIDAN corpus, which only contains Japanese conversations
This decision was made for practical reasons since it is difficult to find spontaneous triadic corpora that contain accurate voice activity information for all speakers.
Dataset selection is important since previous work has shown that VAP performance is dependent on the dataset \cite{sato2024apsipa}.
Future work should also incorporate additional languages like \cite{inoue2024lreccoling} and evaluate multiple datasets for each group size.

\section{Conclusions}\label{sec:conclusion}

This work is the first to incorporate voice activity projection (VAP) into triadic turn-taking prediction.
We trained our triadic turn-taking model using a corpus that included both spontaneous discussion and attentive listening.
All triadic VAP models outperformed the baseline comparison.
These results indicate that our triadic VAP models can accurately predict the next speaker during periods of mutual silence.
Future work will incorporate the triadic VAP model into a spoken dialogue system and evaluate the performance in experimental settings.

\section{Acknowledgements}
This work was supported by JST Moonshot R\&D JPMJPS2011 and JST PREST JPMJPR24I4.
The authors also wish to express their appreciation to members of the speech and audio processing laboratory at Kyoto University for their participation in the data collection.

\bibliographystyle{IEEEtran}
\bibliography{mybib}

\begin{thebibliography}{10}
\providecommand{\url}[1]{#1}
\csname url@samestyle\endcsname
\providecommand{\newblock}{\relax}
\providecommand{\bibinfo}[2]{#2}
\providecommand{\BIBentrySTDinterwordspacing}{\spaceskip=0pt\relax}
\providecommand{\BIBentryALTinterwordstretchfactor}{4}
\providecommand{\BIBentryALTinterwordspacing}{\spaceskip=\fontdimen2\font plus
\BIBentryALTinterwordstretchfactor\fontdimen3\font minus \fontdimen4\font\relax}
\providecommand{\BIBforeignlanguage}[2]{{%
\expandafter\ifx\csname l@#1\endcsname\relax
\typeout{** WARNING: IEEEtran.bst: No hyphenation pattern has been}%
\typeout{** loaded for the language `#1'. Using the pattern for}%
\typeout{** the default language instead.}%
\else
\language=\csname l@#1\endcsname
\fi
#2}}
\providecommand{\BIBdecl}{\relax}
\BIBdecl

\bibitem{sacks1974}
H.~Sacks, E.~A. Schegloff, and G.~Jefferson, ``A simplest systematics for the organization of turn-taking for conversation,'' \emph{Language}, vol.~50, no.~4, pp. 696--735, 1974.

\bibitem{clark2002speechcomm}
H.~H. Clark, ``Speaking in time,'' \emph{Speech Communication}, vol.~36, no.~1, pp. 5--13, 2002, eSCA Workshop on Dialogue and Prosody, September 1999.

\bibitem{stivers2009universals}
T.~Stivers, N.~J. Enfield, P.~Brown, C.~Englert, M.~Hayashi, T.~Heinemann, G.~Hoymann, F.~Rossano, J.~P. De~Ruiter, K.-E. Yoon, and S.~C. Levinson, ``Universals and cultural variation in turn-taking in conversation,'' \emph{Proceedings of the National Academy of Sciences of the United States of America}, vol. 106, no.~26, pp. 10\,587--10\,592, 2009.

\bibitem{bates2003timing}
E.~Bates, S.~D’Amico, T.~Jacobsen, A.~Székely, E.~Andonova, A.~Devescovi, D.~Herron, C.~C. Lu, T.~Pechmann, C.~Pléh, N.~Wicha, K.~Federmeier, I.~Gerdjikova, G.~Gutierrez, D.~Hung, J.~Hsu, G.~Iyer, K.~Kohnert, T.~Mehotcheva, A.~Orozco-Figueroa, A.~Tzeng, and O.~Tzeng, ``Timed picture naming in seven languages,'' \emph{Psychonomic Bulletin \& Review}, vol.~10, no.~2, pp. 344--380, 2003.

\bibitem{roberts2011discourse}
P.~M. Felicia~Roberts and S.~Takano, ``Judgments concerning the valence of inter-turn silence across speakers of american english, italian, and japanese,'' \emph{Discourse Processes}, vol.~48, no.~5, pp. 331--354, 2011.

\bibitem{ward2016aimag}
N.~G. Ward and D.~DeVault, ``Challenges in building highly interactive dialogue systems,'' \emph{AI Magazine}, vol.~37, no.~4, pp. 7--18, 2016.

\bibitem{ward2010interspeech}
N.~G. Ward, O.~Fuentes, and A.~Vega, ``Dialog prediction for a general model of turn-taking,'' in \emph{Interspeech 2010}, 2010, pp. 2662--2665.

\bibitem{skantze2014speechcomm}
G.~Skantze, A.~Hjalmarsson, and C.~Oertel, ``Turn-taking, feedback and joint attention in situated human–robot interaction,'' \emph{Speech Communication}, vol.~65, pp. 50--66, 2014.

\bibitem{ferrer2002icslp}
L.~Ferrer, E.~Shriberg, and A.~Stolcke, ``Is the speaker done yet? faster and more accurate end-of-utterance detection using prosody,'' in \emph{7th International Conference on Spoken Language Processing (ICSLP 2002)}, 2002, pp. 2061--2064.

\bibitem{itoh2009interspeech}
T.~Itoh, N.~Kitaoka, and R.~Nishimura, ``Subjective experiments on influence of response timing in spoken dialogues,'' in \emph{Interspeech 2009}, 2009, pp. 1835--1838.

\bibitem{ekstedt22interspeech}
E.~Ekstedt and G.~Skantze, ``Voice activity projection: Self-supervised learning of turn-taking events,'' in \emph{Interspeech 2022}, 2022, pp. 5190--5194.

\bibitem{inoue2024lreccoling}
K.~Inoue, B.~Jiang, E.~Ekstedt, T.~Kawahara, and G.~Skantze, ``Multilingual turn-taking prediction using voice activity projection,'' in \emph{Proceedings of the 2024 Joint International Conference on Computational Linguistics, Language Resources and Evaluation (LREC-COLING 2024)}, 2024, pp. 11\,873--11\,883.

\bibitem{onishi2023hai}
K.~Onishi, H.~Tanaka, and S.~Nakamura, ``Multimodal voice activity prediction: Turn-taking events detection in expert-novice conversation,'' in \emph{Proceedings of the 11th International Conference on Human-Agent Interaction (HAI '23)}, 2023, p. 13–21.

\bibitem{binger2023icphs}
B.~Jiang, E.~Ekstedt, and G.~Skantze, ``What makes a good pause? investigating the turn-holding effects of fillers,'' in \emph{Proceedings of the 20th International Congress of Phonetic Sciences (ICPhS '23)}, 2023, pp. 3512--3516.

\bibitem{liermann2023emnlp}
W.~Liermann, Y.-H. Park, Y.-S. Choi, and K.~Lee, ``Dialogue act-aided backchannel prediction using multi-task learning,'' in \emph{Findings of the Association for Computational Linguistics: EMNLP 2023}, 2023, pp. 15\,073--15\,079.

\bibitem{chiba2025acl}
Y.~Chiba and R.~Higashinaka, ``Investigating the impact of incremental processing and voice activity projection on spoken dialogue systems,'' in \emph{Proceedings of the 31st International Conference on Computational Linguistics}, 2025, pp. 3687--3696.

\bibitem{ekstedtphd}
E.~Ekstedt, ``Predictive modeling of turn-taking in spoken dialogue : Computational approaches for the analysis of turn-taking in humans and spoken dialogue systems,'' Ph.D. dissertation, KTH, Speech, Music and Hearing, TMH, 2023, qC 20231115.

\bibitem{librispeech}
V.~Panayotov, G.~Chen, D.~Povey, and S.~Khudanpur, ``Librispeech: An asr corpus based on public domain audio books,'' in \emph{2015 IEEE International Conference on Acoustics, Speech and Signal Processing (ICASSP)}, 2015, pp. 5206--5210.

\bibitem{riviere2020icassp}
M.~Rivière, A.~Joulin, P.-E. Mazaré, and E.~Dupoux, ``Unsupervised pretraining transfers well across languages,'' in \emph{ICASSP 2020 - 2020 IEEE International Conference on Acoustics, Speech and Signal Processing (ICASSP)}, 2020, pp. 7414--7418.

\bibitem{vaswani2017}
A.~Vaswani, N.~Shazeer, N.~Parmar, J.~Uszkoreit, L.~Jones, A.~N. Gomez, L.~u. Kaiser, and I.~Polosukhin, ``Attention is all you need,'' in \emph{Advances in Neural Information Processing Systems}, vol.~30, 2017.

\bibitem{pytorch}
A.~Paszke, S.~Gross, F.~Massa, A.~Lerer, J.~Bradbury, G.~Chanan, T.~Killeen, Z.~Lin, N.~Gimelshein, L.~Antiga, A.~Desmaison, A.~Kopf, E.~Yang, Z.~DeVito, M.~Raison, A.~Tejani, S.~Chilamkurthy, B.~Steiner, L.~Fang, J.~Bai, and S.~Chintala, ``Pytorch: An imperative style, high-performance deep learning library,'' in \emph{Advances in Neural Information Processing Systems}, vol.~32, 2019.

\bibitem{yuan2007icphs}
J.~Yuan, M.~Liberman, and C.~Cieri, ``Towards an integrated understanding of speech overlaps in conversation,'' in \emph{Proceedings of the 16th International Congress of Phonetic Sciences (ICPhS '07)}, 2007, pp. 1337--1340.

\bibitem{duncan1972signals}
S.~Duncan, ``Some signals and rules for taking speaking turns in conversations,'' \emph{Journal of Personality and Social Psychology}, vol.~23, no.~2, pp. 283--292, 1972.

\bibitem{gravano2011computer}
A.~Gravano and J.~Hirschberg, ``Turn-taking cues in task-oriented dialogue,'' \emph{Computer Speech \& Language}, vol.~25, no.~3, pp. 601--634, 2011.

\bibitem{aldeneh2018icassp}
Z.~Aldeneh, D.~Dimitriadis, and E.~M. Provost, ``Improving end-of-turn detection in spoken dialogues by detecting speaker intentions as a secondary task,'' in \emph{2018 IEEE International Conference on Acoustics, Speech and Signal Processing (ICASSP)}, 2018, pp. 6159--6163.

\bibitem{sato2024apsipa}
Y.~Sato, Y.~Chiba, and R.~Higashinaka, ``Investigating the language independence of voice activity projection models through standardization of speech segmentation labels,'' in \emph{2024 Asia Pacific Signal and Information Processing Association Annual Summit and Conference (APSIPA ASC)}, 2024, pp. 1--6.

\end{thebibliography}

\end{document}